\pdfoutput=1
\def\year{2022}\relax
\documentclass[letterpaper]{article} %
\usepackage{aaai22}  %
\usepackage{times}  %
\usepackage{helvet}  %
\usepackage{courier}  %
\usepackage[hyphens]{url}  %
\usepackage{graphicx} %
\usepackage{natbib}  %
\usepackage{caption} %
\DeclareCaptionStyle{ruled}{labelfont=normalfont,labelsep=colon,strut=off} %
\usepackage{algorithm}

\usepackage{newfloat}
\usepackage{listings}
\floatstyle{ruled}
\newfloat{listing}{tb}{lst}{}
\floatname{listing}{Listing}
\usepackage{amsmath}
\usepackage{acro}
\usepackage[noend]{algpseudocode}
\usepackage{booktabs}
\usepackage{soul}
\usepackage{subcaption}
\usepackage{xspace}

\usepackage[capitalise]{cleveref}

\DeclareAcronym{CNN}{short=CNN,long=convolutional neural network}
\DeclareAcronym{DSE}{short=DSE,long=design space exploration}
\DeclareAcronym{FL}{short=FL,long=federated learning,short-indefinite=an}
\DeclareAcronym{LUT}{short=LUT,long=lookup table}
\DeclareAcronym{MAC}{short=MAC,long=multiply-accumulate operation}
\DeclareAcronym{ML}{short=ML,long=machine learning,short-indefinite=an}
\DeclareAcronym{NN}{short=NN,long=neural network,short-indefinite=an}
\DeclareAcronym{SGD}{short=SGD,long=stochastic gradient descent}

\newcommand{\rpm}{\raisebox{.2ex}{$\scriptstyle\pm$}}

\newcommand{\ourtech}{\mbox{\emph{DISTREAL}}\xspace}
\title{
	DISTREAL: Distributed Resource-Aware Learning in Heterogeneous Systems
}
\author{
	Martin Rapp\textsuperscript{\rm 1},
	Ramin Khalili\textsuperscript{\rm 2},
	Kilian Pfeiffer\textsuperscript{\rm 1},
	J\"org Henkel\textsuperscript{\rm 1}
}
\affiliations {
	\textsuperscript{\rm 1} Karlsruhe Institute of Technology, Karlsruhe, Germany\\
	\textsuperscript{\rm 2} Huawei Research Center, Munich, Germany\\
	martin.rapp@kit.edu, ramin.khalili@huawei.com, kilian.pfeiffer@kit.edu, henkel@kit.edu
}

\begin{document}
	
\maketitle

\begin{abstract}
We study the problem of distributed training of \acp{NN} on devices with heterogeneous, limited, and time-varying availability of computational resources.
We present an adaptive, resource-aware, on-device learning mechanism, \ourtech, which is able to fully and efficiently utilize the available resources on devices in a distributed manner, increasing the convergence speed.
This is achieved with a dropout mechanism that dynamically adjusts the computational complexity of training \iac{NN} by randomly dropping filters of convolutional layers of the model.
Our main contribution is the introduction of \iac{DSE} technique, which finds Pareto-optimal per-layer dropout vectors with respect to resource requirements and convergence speed of the training.
Applying this technique, each device is able to dynamically select the dropout vector that fits its available resource without requiring any assistance from the server.
We implement our solution in \iac{FL} system, where the availability of computational resources varies both between devices and over time, and show through extensive evaluation that we are able to significantly increase the convergence speed over the state of the art without compromising on the final accuracy.
\end{abstract}

\section{Introduction}
\label{sec:intro}

Deep learning has achieved impressive results in a number of diverse domains, such as image classification~\citep{mobilenet,efficientnet}, board and video games~\citep{2016-nature-go,a3c}, and is widely applied to distributed systems, such as mobile and sensor networks~\citep{2019-cst-mobile}, as we consider in this paper.
Centralized deep learning techniques, where the training is performed in a single location (e.g., a data center), is often costly, as data would need to be collected and sent all over the network to that centralized entity~\citep{shi2020communication}, and might not be feasible/authorized if the training uses users' private data.  
\ac{FL}~\citep{federated} emerged as an alternative to such techniques, performing distributed learning on each device with the locally available data. 

\ac{FL} has proven effective in large-scale systems~\citep{bonawitz2019federated, liu2019lifelong, chen2020fedhealth}.
However, training of a deep \ac{NN} model is resource-hungry in terms of computation, energy, time, etc.\ \citep{nns_are_expensive}, and it is rather unrealistic to assume that all devices in an FL system can perform all types of training computations all the time, especially if the training is distributed on edge devices, e.g., as suggested for 6G systems.\footnote{EU's Horizon Europe~\cite{DATA02, DATA03} calls for proposals, as well as 5GIA (5G Infrastructure Association) and SRIA (Strategic Research and Innovation Agenda) reports~\cite{5gIA,SRIA} detail such a vision.}
This is as the computational capabilities of devices participating in \iac{FL} system may be heterogeneous, e.g., different hardware, different generations~\citep{bonawitz2019federated}.
\emph{More importantly, the resources available on a device for training could change over time}.
This could for instance be due to shared resource contention~\citep{dhar2019device}, where CPU time, cache memory,
energy, etc.\ are shared between the learning and parallel tasks.
We illustrate this with the following two examples.

1)~Edge computing has been employed in ML-based real-time video analytics, where each edge device processes images from several camera modules \cite{ananthanarayanan2017real}.
Currently, edge devices mostly perform inference, but there is a clear trend towards additionally performing distributed learning via \ac{FL} \cite{zhou2019edge}.
The learning task shares computational resources
with the inference tasks.
The inference workload depends on the activity in the video images and changes over time, as processing is skipped for subsequent similar images to save resources \cite{ananthanarayanan2017real}.
These changes happen fast, i.e., in the order of seconds \cite{zhang2017live}, while \ac{FL} round times may be minutes \cite{bonawitz2019federated}.
2)~Google \emph{GBoard} \cite{yang2018applied} trains a next-word-prediction model using \ac{FL} on end users' mobile phones.
To avoid slowing down user applications, and thereby degrading the user experience, training is performed only when the device is charging and idle, and aborted when these conditions change.
This introduces a bias towards certain devices and users, degrading the model accuracy \cite{yang2018applied}.
This can be resolved by allowing training also when the device is in use, but only using free resources. %
Smartphone workloads change within seconds \cite{tu2014performance}, which is faster than the \emph{GBoard} round time of several minutes. %
In both examples, the learning task is subject to fast-changing resource availability.

While several works study the problem of heterogeneity across devices~\citep{Li2020,imteaj2020federated}, time-varying resource availability has so far been neglected.
In this paper, we propose a distributed, resource-aware, adaptive, on-device learning technique, \ourtech, which enables us to fully exploit and efficiently utilize available resources on devices for the training, dealing with all these types of heterogeneity.
Our objective is to maximize the accuracy that is reached after limited training time on devices, i.e., convergence speed.
To fulfill this goal, we should make sure that C1) The available resources on a device are \emph{fully} exploited.
This requires fine-grained adjustability of the training model on a device, and a method to instantly react to changes; and C2) The available, limited, resources on a device are used \emph{efficiently}, to maximize the accuracy improvement and hence the overall convergence speed.
Specifically, we provide the following novel contributions:
\begin{itemize}
    \item We introduce and formulate the problem of heterogeneous time-varying resource availability in \ac{FL}.
    \item We propose a dropout technique to adjust the computational complexity (resource requirements) of training the model  \emph{at any time}.
    Thereby, each device locally decides the dropout setting which fits its available resources, without requiring any assistance from the server, addressing C1.
    This is different from the state-of-the-art techniques, such as \citep{caldas2018expanding,horvath2021fjord,xu2019elfish,diao2020heterofl}, where the server is responsible for regulating resource requirements of training for each device at the beginning of each training round, which may take several minutes. %
    \item We show that using different per-layer dropout rates achieves a much better trade-off between the resource requirements and the convergence speed, compared to using the same rate at all layers as the state of the art~\citep{caldas2018expanding,diao2020heterofl}, addressing C2.
    We present \iac{DSE} technique to automatically find the Pareto-optimal dropout vectors at design time.
\end{itemize}

We implement our solution \ourtech in \iac{FL} system, in which the availability of computational resources varies both between devices and over time.
We show through extensive evaluation that \ourtech significantly increases the convergence speed over the state of the art, and is robust to the rapid changes in resource availability at devices, without compromising on the final accuracy.

\section{System Model and Problem Definition}
\label{sec:model}

\paragraph{System Model}
We target a distributed system, which comprises one \emph{server} and $N$ distributed \emph{devices} that act as clients.
Each device~$i$ holds its own local training data~$X_i$.
The system uses \ac{FL} for decentralized training of \iac{NN} model from the distributed data.
We target a synchronous coordination scheme, which divides the training into many \emph{rounds}.
At the beginning of a round, the server selects $n$ devices to participate in the training.
Each selected device downloads the recent model from the server, trains it with its local data, and sends weight updates back to the server.
The server combines all received weight updates to a single update by weighted averaging.
Updates from devices that take too long to perform the training (stragglers)
are discarded.

\paragraph{Device Resource Model}
The devices are subject to time-varying limited computational resource availability for training.
To which degree the availability of a certain resource affects the training time of \iac{NN} depends on the \ac{NN} and hyperparameters, but also on the deep learning library implementation and the underlying hardware~\citep{nn_acceleration}.
We abstract from such specifics of the hardware and software implementation, and from the constrained physical resource to keep this work applicable to many systems
by representing the resource availability in the number of \acp{MAC} that a device can calculate per time given its specifications and available resources.
\ac{MAC} operations are the fundamental building block of \acp{NN} (e.g., fully-connected and convolutional layers) and account for the great majority of operations~\citep{alexnet}.
In the appendix, we also provide experimental evidence for the suitability of \acp{MAC}/s as an abstract metric.
Resource availability varies between devices and over time.
Therefore, these resource availabilities~$r_i(t)$ depend on the device~$i$, and the current time~$t$.
Resources may change at any time, i.e., also within \iac{FL} round.
Resources are not required to be known ahead of time.

\paragraph{Objective}
Our objective is to maximize the convergence speed
 of training%
, i.e., the reached accuracy after a number of rounds, under heterogeneous (between devices and over time) resource availability.

\section{Related Work}
\label{sec:related}

Many works on resource-aware machine learning focus on resource-aware \emph{inference}~\citep{tann2016runtime,yu2018slimmable,amir2018priority,li2021dynamic}.
These techniques allow adapting the inference to dynamically changing availability of resources at run time but are not applicable to training.
Resource-aware \emph{training} is recently getting increasing attention, mostly in the context of \ac{FL}.
Most attention has so far been paid to limited \emph{communication} resources, leading to solutions, such as compression, quantization, and sketching~\citep{shi2020communication,thakker2019compressing}.
Importantly, these works do not reduce the \emph{computational} resources for training, as they are applied after local training has finished.
These works are complementary/orthogonal to our work and can be adopted to our solution (see the section on the run-time technique).
Techniques on computation-resource-aware training
can be categorized into two classes: techniques that always train the full \ac{NN}
on each device
but with fewer data/relaxed timing and techniques that train subsets of the \ac{NN}.

\paragraph{Train Full \acp{NN}}
FedProx~\citep{li2020federated} allows devices participating in \iac{FL} system to deliver partial results to the server by dropping training examples that could not be processed with the available resources. %
Our previous work \cite{rapp2020distributed} studied multi-head networks where each device uses the head that fits its available resources.
Devices only synchronize the weights of the first shared layers.
However, this technique has low adaptability as only a few resource levels can be supported.
Asynchronous variants of \ac{FL} have been proposed that allow devices to finish training at any time~\citep{chen2019asynchronous,xie2020asynchronous}.
However, asynchronous synchronization may reduce the convergence stability~\citep{federated,xu2019elfish}.
Techniques based on Federated Distillation~\citep{Li2019,chang2019cronus,lin2020ensemble} synchronize knowledge between devices by exchanging labels on a public dataset instead of exchanging \ac{NN} weights. %
Therefore, each device has the design flexibility to use \iac{NN} model according to its constraints.
However, Federated Distillation
cannot cope with time-varying resources.

\paragraph{Train \ac{NN} Subsets}
Several techniques perform training only on a dynamic subset of the \ac{NN}, to be able to fit the resource requirements of training to the resource availability on each device. %
FjORD~\citep{horvath2021fjord}, \citet{Yu2021Federated}, and HeteroFL~\citep{diao2020heterofl} select subsets of the \ac{NN} for each device at the beginning of each round. %
They select the subsets in a hierarchical way, where smaller subsets are fully contained in larger subsets.
HeteroFL introduces a shrinkage ratio~$s$ that defines the ratio of removed hidden channels to reduce the resource requirements of the \ac{NN}.
The same parameter~$s$ is applied repeatedly to all layers to obtain several subsets with decreasing resource requirements.
Using hierarchical subsets restricts the granularity of resource requirements, as increasing the number of supported subsets reduces the achievable accuracy~\citep{tann2016runtime}.
This limitation can be avoided by selecting subsets randomly, i.e., use different subsets in every round.
ELFISH \cite{xu2019elfish} randomly removes neurons before training on slow devices at the beginning of a round.
\citet{graham2015efficient} study the suitability of dropout~\citep{dropout} to reduce resource requirements.
They find that computations can only be saved if dropout is done in a structured way, i.e., the same neurons are dropped for all samples of a mini-batch.
Federated Dropout~\citep{caldas2018expanding} has been originally proposed to reduce the communication and computation overhead of \ac{FL}.
They perform dropout at the server and train a repacked smaller network on the devices.
The dropout masks are changed randomly in each round, which results in all parts of the \ac{NN} being trained eventually.
However, they use the same dropout rate for all devices and a single dropout rate for all layers.
All these works select the trained subset at the server, which may reduce the communication volume, but importantly does not allow to adapt to changing resource availability on the devices within a round.

\section{Resource-Aware Training of NNs}
\label{sec:technique}

Our technique
comprises two parts.
At run time (online), we dynamically drop parts of the \ac{NN} using an adapted version of dropout~\citep{dropout}.
The Pareto-optimal vectors of dropout rates w.r.t.\ convergence speed and resource requirements are obtained at design time (offline) using \iac{DSE}.
Before going into the details of our contribution, we introduce dropout as the basis of our technique.

\paragraph{Dropout to Reduce Computations In Training}
Dropout was originally designed as a regularization method to mitigate overfitting~\citep{dropout}.
It randomly drops individual neurons during training with a certain probability (dropout rate).
This results in an irregular fine-grained pattern of dropped neurons.
All major deep learning libraries perform dropout by calculating the output of all neurons and multiplying the dropped ones with~$0$ \citep{tensorflow,pytorch}.
This wastes computational resources; it would be more efficient to not calculate values that are going to be dropped.
However, convolutional and fully-connected layers are implemented as matrix-vector or matrix-matrix operations that are heavily optimized with the help of vectorization~\citep{tensorflow,pytorch}.
Skipping the calculation of individual values results in sparse matrix operations, which breaks vectorization, increasing the required resources instead of decreasing them~\citep{song2019approximate}.

\begin{figure}
    \centering
	\includegraphics{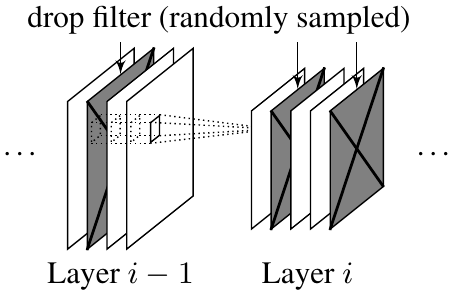}
	\caption{
	    Filter-based structured dropout in a convolutional layer maintains regularity in the calculations while significantly reducing the required computations.
	}
	\label{fig:skip_filters}
\end{figure}

To reduce the number of computations, the dropout pattern needs to show some regularity that still allows using vectorization of dense matrix operations.
This can be achieved by dropping contiguous parts of the computation~\citep{graham2015efficient}.
Modern \acp{NN} consist of many different layer types such as convolutional, pooling, fully-connected, activation, or normalization layers.
Many of these layers are computationally lightweight (e.g., pooling), while some contain the majority of computations (convolutional and fully-connected layers).
In state-of-the-art convolutional \acp{NN}, the convolutional part requires orders of magnitude more \acp{MAC} than the fully-connected part. (See the appendix for an experimental analysis.)
We, therefore, argue that a technique to save computations needs to target convolutional layers.
\cref{fig:skip_filters} depicts filter-based structured dropout in a convolutional layer, as we apply in this paper: instead of dropping individual pixels in the output, whole filters are dropped stochastically.
This approach reduces the number of computations while allowing to keep existing vectorization methods.

\begin{figure}
    \centering
	\includegraphics{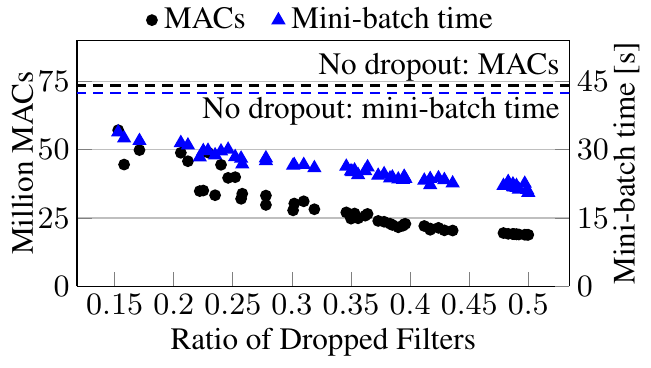}
	\caption{
	    The number of \acp{MAC} and mini-batch training time decrease quadratically with the ratio of dropped filters.
	}
	\label{fig:sd_performance}
\end{figure}

\cref{fig:sd_performance} depicts how the number of \acp{MAC} of the forward pass evolves when we apply different vectors of per-layer dropouts for DenseNet-40 (details in the experimental evaluation).
We apply the DSE technique introduced in this paper to determine these vectors and show in the x-axis the resulting ratio of dropped filters\footnote{Multiple vectors may results in the same ratio of dropped filters, while providing different convergence / resource requirement trade-offs, explaining why for some ratios we have multiple \acp{MAC}.}.
We observe that the number of \acp{MAC} decreases almost quadratically with this ratio.
We also report the training time of a single mini-batch on a Raspberry Pi 4, which serves as an example for an IoT device, using an implementation of this structured dropout in PyTorch~\citep{pytorch}, which is publicly available\footnote{\url{https://git.scc.kit.edu/CES/DISTREAL}}.
A training step (forward pass, backpropagation, and weight updates) requires about ${\sim}2{\times}$ more \acp{MAC} than the forward pass alone~\citep{ai_and_compute}.
The mini-batch training time shows a similar trend as the number of \acp{MAC} but with an offset.
This is because our implementation does not modify the backend of PyTorch to be aware of this dropout, which results in copy operations of weight tensors to repack them.
As a consequence, the measured benefits are smaller than what is theoretically achievable.
Changing the backend would get the benefits closer to the optimum but is not easily doable because of closed sources (e.g., of CUDA).
In summary, this experiment shows that structured dropout significantly reduces the required computational resources.

The dropout rates also change the convergence speed.
A higher dropout rate in a layer means that in each training update, a smaller fraction of the layer's weights is updated, thereby slowing down the training.
As a consequence, the dropout rate determines a trade-off between the resource requirements and the convergence speed.
The dropout rate should always be selected as low as the available resources allow.
To cope with changing resource availability, we propose to \emph{dynamically change the dropout rate at run time}.
Inference uses always the whole model.

\paragraph{Design-Time DSE: Find Pareto-Optimal Dropout Vectors}
\label{sec:dse}

The resource requirements (\acp{MAC}) and convergence speed both depend on the dropout rates of \emph{each layer}.
Prior works restrict themselves to choosing a common dropout rate for all layers~\citep{caldas2018expanding}.
Relaxing this restriction opens up a larger design space, where each dropout rate of each layer is adjusted towards a better trade-off between resource requirements and training convergence.
However, this design space could be too large to be explored manually.
For example, DenseNet-100 has 99 convolutional layers that each need to be assigned a dropout rate.
Some works apply simple parametric functions of the depth to similar problems \cite{huang2016deep}.
However, this only works in case of a homogeneous \ac{NN} structure, where properties of layers (e.g., \acp{MAC}) change monotonically.
For instance, DenseNet layers alternate between computationally lightweight and complex, rendering a simple parametric function sub-optimal.
This section describes the required automated \ac{DSE} technique to efficiently explore such a large space.
The DSE is executed only once at design time (offline).

Specifically, the design space contains all combinations of dropout values per layer.
We select dropout values from the continuous range $[0,0.5]$ because higher values reduce the final achievable accuracy, as we observe in our experiments, as well as indicated in previous studies~\citep{dropout}.
For \iac{NN} with $k$ convolutional layers, the design space is $[0,0.5]^k$.
We have two objectives, the resource requirements and the convergence speed.

\emph{Resource Requirements:} As discussed in the device resource model, the number of \acp{MAC} is an implementation-independent representation of the resource requirements.
Dropout is a probabilistic process, i.e., the number of \acp{MAC} varies between different update steps.
The resource requirements with a certain dropout vector is represented by the expectation value of the number of \acp{MAC} of the forward pass.
This number can be analytically computed depending on the layer topology, the dropout rate of this layer and preceding layers.
The appendix lists equations for different layer types.

\emph{Convergence Speed:} The convergence speed with a certain dropout vector is measured by observing the accuracy change when training.
Exploring the search space takes too long if a full training with every candidate dropout vector is performed.
Instead, we assess the accuracy change after a short training, similar to learning curve extrapolation
in neural architecture search~\citep{baker2017accelerating}.
We train for 64 mini-batches with batch size 64, which allows us to explore many candidate dropout vectors in a reasonable time.
This corresponds to the amount of data collected by very few devices.
To reduce the impact of random initialization, the \ac{NN} is not trained from scratch but from a snapshot after partially training it on a distorted version of the dataset.
For instance, we reduced the brightness, contrast, and saturation to 0.5 of the original value for CIFAR-10/100 datasets.
The DSE, therefore, does not require access to the devices' data, but only access to a small amount of similar (or even synthetic) data.
To further reduce the impact of random variations, we repeat this with three different random seeds.
The convergence speed is represented by the average accuracy improvement.

\begin{figure}
    \centering
    \includegraphics{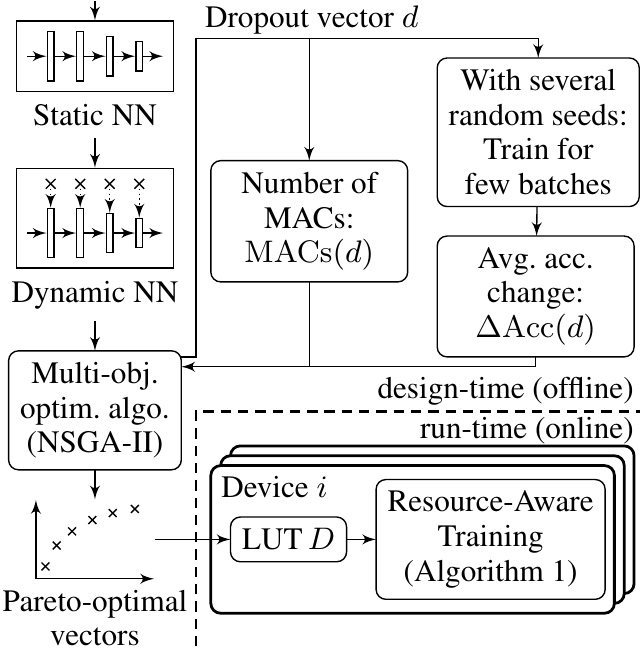}
    \caption{
        Efficient resource-aware training comprises the \ac{DSE} to find Pareto-optimal vectors of dropout rates per layer
        and resource-aware training on each device at run time.
    }
    \label{fig:dse_flow}
\end{figure}

\cref{fig:dse_flow} shows our \ac{DSE} flow.
The problem of finding Pareto-optimal dropout vectors is a multi-objective optimization.
This is a well-studied class of problems with many established algorithms.
Evolutionary algorithms have successfully been employed for neural architecture search~\citep{nas}, which is related to the problem studied in this section.
Note that we are not searching for an architecture, but tune parameters of a given architecture. 
The output of the \ac{DSE} is the Pareto-front of dropout vectors.
To have a large variety of options to chose from at run time, but also keep a low number of vectors to be stored, the Pareto-front should be approximately equidistantly represented.
We use the NSGA-II~\citep{nsga2} genetic algorithm from the pygmo2 library~\citep{pagmo}.
NSGA-II explores the search space by crossover (combining parts of two dropout vectors) and mutation (random changes) of good dropout vectors w.r.t.\ the objective function, and is designed to obtain dropout vectors that are equidistantly distributed across the Pareto-front.
Thereby, an \emph{individual} is one dropout vector containing the per-layer dropout rates.
For our largest studied \ac{NN}, DenseNet-100, this is 99 float values between 0 and 0.5.
A \emph{population} is a set of individuals.
We use a population size of 64.
A \emph{generation} performs one optimization step on the population with the goal to find the Pareto-front.
The optimization minimizes the following two-dimensional \emph{fitness} function~$f(d)$ for a dropout vector~$d$,
which normalizes the values of the resource requirements~$\mathrm{MACs}(d)$ and convergence speed~$\Delta\mathrm{Acc}(d)$ to the range $[0,1]$:
\begin{equation}
\renewcommand*{\arraystretch}{1.8}
f(d) = \begin{pmatrix}
\frac{
    \mathrm{MACs}(d)-\mathrm{MACs}(\{0.5,...,0.5\})
}{
    \mathrm{MACs}(\{0,...,0\})-\mathrm{MACs}(\{0.5,...,0.5\})
}\\
\frac{
    \Delta\mathrm{Acc}(\{0,...,0\})-\Delta\mathrm{Acc}(d)
}{
    \Delta\mathrm{Acc}(\{0,...,0\})-\Delta\mathrm{Acc}(\{0.5,...,0.5\})
}
\end{pmatrix}
\label{eq:fitness}
\end{equation}

\begin{figure}
	\centering
	\includegraphics{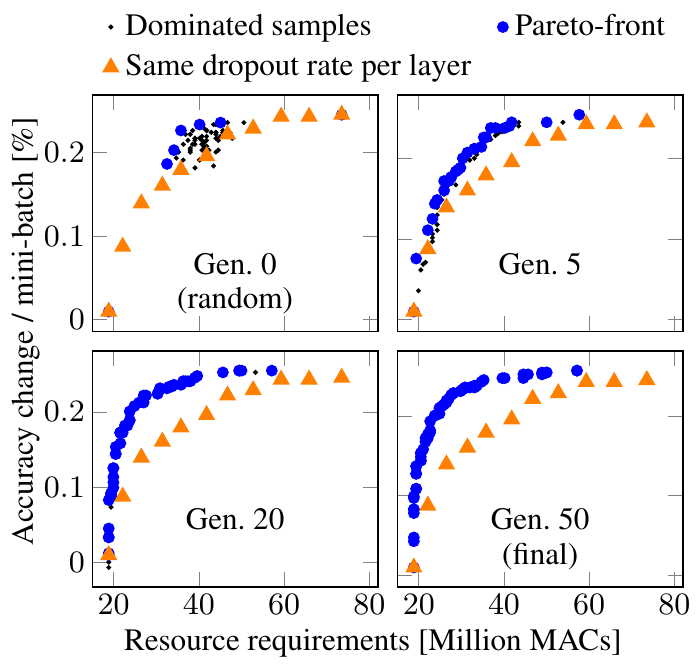}
    \caption{
        The evolving Pareto-front for DenseNet-40 significantly outperforms setting the same rate for all layers.
    }
    \label{fig:dse_populations}
\end{figure}

\cref{fig:dse_populations} shows the evolving population of dropout vectors for DenseNet-40. %
The initial population comprises random dropout vectors.
We add two samples to the initial population (all dropout values are 0 / 0.5) to accelerate the exploration of the Pareto-front (leverage the crossover operation).
After 50 generations of NSGA-II, the Pareto-front has fully evolved and shows a continuous trade-off between resource requirements and convergence speed.
Importantly, the Pareto-front found by \ac{DSE} provides a significantly better trade-off between resource requirements and convergence speed compared to using the same dropout rate for all layers.

\paragraph{Run Time: Resource-Aware Training of \acp{NN}}

After finding the Pareto-optimal dropout vectors, they are stored in a \ac{LUT}~$D$, along with the corresponding number of \acp{MAC}.
The \ac{LUT} is small in size (e.g., 25\,kB for DenseNet-100 for storing 64 dropout vector of 99 dropout values and the number of \acp{MAC}, each in 32-bit format) and stays constant for all rounds.
At run time, a device selects the dropout vector~$d$ that best corresponds to its resource availability.
If resource availability changes at the device, the dropout vectors can be adjusted to these changes at almost zero overhead before every mini-batch.
No weight copies, recompilation, repacking of weights, etc.\ are required~for adapting the resource requirements.

\begin{algorithm}[t]
	\caption{Each Selected Device~$i$ (Client)}
	\footnotesize
	\label{algo:device}
	\begin{algorithmic}
	    \Require $D$: LUT of Pareto-optimal dropout vectors (from DSE)
		\State receive $\theta_{init}$ from server
		\State $\theta\gets\theta_{init}, c\gets 0$
		\For{\textbf{each} $b \in X_i$} \Comment{\emph{iterate over mini-batches from local data}}
		    \State{$r\gets r_i(t)$} \Comment{\emph{current resource availability}}
		    \State{$d\gets D[r]$} \Comment{\emph{resource-aware dropout vector}}
		    \State{Update dropout values of local NN with $d$}
		    \State $\theta \gets \theta - \eta \frac{\partial}{\partial\theta} L(b;\theta)$ \Comment{\emph{update step}}
		    \State{$c\gets c+\mathrm{MACs}(d)$} \Comment{\emph{accumulate computations}}
		\EndFor
		\State{send $(\theta {-} \theta_{init},c)$ to server} \Comment{\emph{weight update and computations}}
	\end{algorithmic}
\end{algorithm}

\begin{algorithm}[t]
	\caption{Server}
	\footnotesize
	\label{algo:server}
	\begin{algorithmic}
		\State $\theta_{0} \gets$ random initialization
		\For{\textbf{each} round~$t=1,2,\ldots$}
		    \State $K\gets$select $n$ devices
			\State broadcast $\theta_{t-1}$ to selected devices~$K$
			\State receive $(d\theta_i,c_i)$ from devices $i\in K$
			\State $C \gets \sum_{i\in K} c_i$ \Comment{\emph{tot. computations}}
			\State $d\Theta \gets \sum_{i\in K} c_i {\cdot} d\theta_i$ \Comment{\emph{weighted sum}}
			\State $d\theta \gets d\Theta /C$ \Comment{\emph{weighted average}}
			\State $\theta_{t} \gets \theta_{t-1} + d\theta$
		\EndFor
	\end{algorithmic}
\end{algorithm}

In \iac{FL} setting, each device selects its dropout vector at run time according to its resource availability, as shown in \cref{algo:device}.
This is done at the granularity of single mini-batches, i.e., devices can quickly react to changes.
Additionally, the server does not need to know the resource availability at each device at the beginning of the round, reducing signaling overhead, and avoiding the requirement to know resource availability ahead of time.
This is important to maintain scalability with the number of devices.
At the end of each round, the devices report back the weight updates and the computational resources they put into training (number of \acp{MAC}, as stored in the \ac{LUT}). %
The server (\cref{algo:server}) performs a weighted averaging of the received updates w.r.t.\ the devices' reported computational resources.
Thereby, updates from devices that have trained with lower dropout rates, are weighted stronger.
This is an extension of \emph{FedAvg}~\citep{federated}, which performs weighted averaging only based on the number of mini-batches.
In the case of constant and same resource availability on all devices, our coordination technique behaves the same as \emph{FedAvg}.
As we do not change the type of data exchanged between the devices and the server, compared to \emph{FedAvg}, we can still apply and adopt techniques that mitigate communication aspects, such as compression and sketched updates~\citep{shi2020communication}.

\section{Experimental Results}
\label{sec:experiments}

This section demonstrates the benefits of \ourtech with heterogeneous resource availability in \iac{FL} system.

\paragraph{Experimental Setup}
\label{sec:experiments:setup}

We study
synchronous \ac{FL} as described in the system model.
We report the classification accuracy of the synchronized model at the end of each round.
Our main performance metric is the \emph{convergence speed}, i.e.,
the accuracy achieved after a certain number of rounds,
but we also report the final accuracy after convergence.

\begin{table}
    \small
    \centering
    \begin{tabular}{cccc}
		\toprule
		& FEMNIST & CIFAR-10 & CIFAR-100 \\
		\midrule
		\#Devices & $3{,}550$ & $100$ &  $100$ \\
		\#Samples/device & $181{\pm}70.7$ & $500$ & $500$ \\
		Devices/round & 35 & 10 & 10 \\
		Resources var. & 3$\times$ & 4$\times$ & 4$\times$ \\
		\bottomrule
	\end{tabular}
    \caption{System configuration for \ac{FL}.}
    \label{tab:fl_setting}
\end{table}

The three datasets used in our experiments are \emph{Federated Extended MNIST} (FEMNIST) \citep{cohen2017emnist} with non-independently and identically distributed (non-iid) split data, similar to LEAF \citep{caldas1812leaf}, and \mbox{CIFAR-10/100} \citep{cifar10}.
FEMNIST consists of 641,828 training and 160,129 test examples, each a $28{\times}28$ grayscale image of one out of 62 classes (10 digits, 26 upper- and 26 lower-case letters).
CIFAR-10 consists of 50,000 training and 10,000 test examples, each a $32{\times}32$ RGB image of one out of 10 classes such as airplane or frogs.
CIFAR-100 is similar to CIFAR-10 but uses 100 classes.
\cref{tab:fl_setting} summarizes the configurations.

For FEMNIST, we use a similar network as used in Federated Dropout \citep{caldas2018expanding}, with a depth of 4~layers, requiring 4.0~million \acp{MAC} in the forward pass. We use DenseNet \citep{densenet} for CIFAR-10 and CIFAR-100 with growth rate $k=12$ and depth of $40$ and $100$, respectively.
This results in 74~million \acp{MAC} for CIFAR-10 and 291~million \acp{MAC} for CIFAR-100 in the forward pass.
The \ac{DSE} for these \acp{NN} takes around 15, 270, and 330 compute-hours, respectively, on a system with an Intel Core i5-4570 and an NVIDIA GeForce GTX 980.
More details about the \ac{NN} configurations and the computational complexity of the \ac{DSE} are presented in the appendix.

We compare \ourtech to four baselines:
\begin{enumerate}
    \item \textbf{Full resource availability.} All devices have the full resources to train the full \ac{NN} in each round. This is a theoretical baseline, which serves as an upper bound.
    \item \textbf{Small network.} The \ac{NN} complexity is reduced to fit the weakest device. Thereby, each device can train the full (reduced) \ac{NN} in each round with FedAvg. For CIFAR-10 and CIFAR-100, we reduce the depth of DenseNet to 19 and 40, respectively. Because the network of FEMNIST already has only a few layers, we reduce the number of filters of the convolutional layers.
    \item \textbf{Federated Dropout} as in \citep{caldas2018expanding}. Similar to our technique, it uses dropout to reduce the computational complexity. However, the same dropout rate is used for all layers. To have a fair comparison, we extend the technique of \citet{caldas2018expanding} to allow for different dropout rates for different devices according to the resource availability. The rates are determined by the server at the start of each round as in the original technique.
    \item \textbf{HeteroFL} as in \citep{diao2020heterofl}. It uses a shrinking ratio $0{<}s{<}1$. The \ac{NN} is divided into several levels~$p=1, 2, \ldots$, where level $p$ reduces the width of each hidden channel %
    to a fraction $s^{p-1}$. This is done on the server at the beginning of each round. 
    \citep{diao2020heterofl} provides no details on how to set $s$. We use $s{=}0.7$, as it shows the best performance.
\end{enumerate}

\begin{figure}[t]
    \centering
    \begin{subfigure}{\linewidth}
        \centering
	    \includegraphics{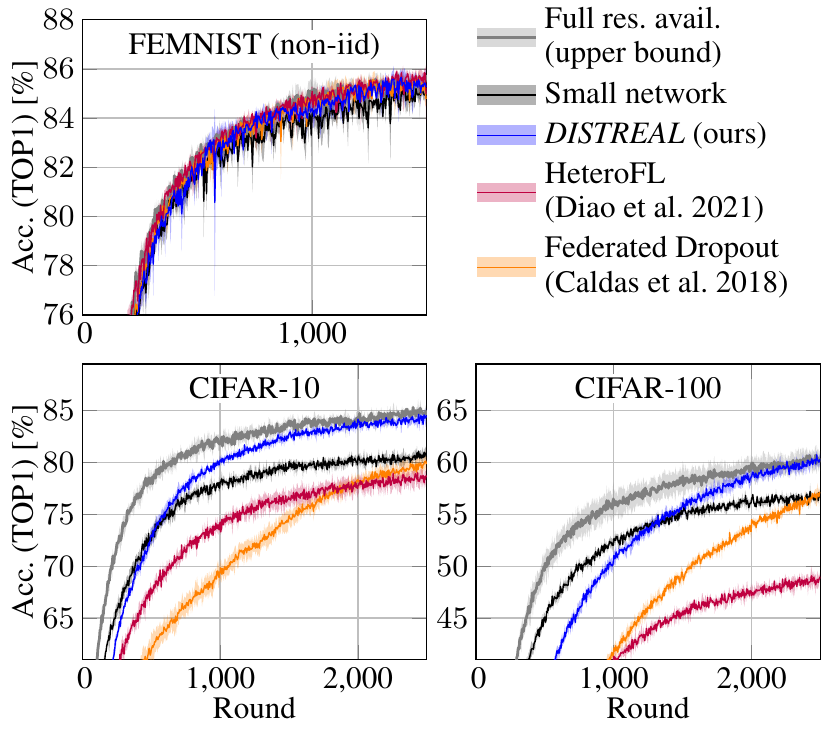}
        \caption{Convergence speed}
        \label{fig:results_fl:convergence}
    \end{subfigure}
    \begin{subfigure}{\linewidth}
    	\centering
        \small
    	\vspace{2mm}
    	\begin{tabular}{p{2.0cm}|ccc}
    		\toprule
    		& FEMNIST & CIFAR-10 & CIFAR-100 \\
    		\midrule
    		Full res.\ avail.\ (upper bound) &
    		    $87.4 {\rpm} 0.3$ &
    		    $87.8 {\rpm} 0.1$ &
    		    $65.6 {\rpm} 0.5$ \\
    		\midrule
    		Small NN &
    		    $86.4 {\rpm} 0.1$ &
    		    $82.2 {\rpm} 0.6$ &
    		    $57.8 {\rpm} 0.5$ \\
    		DISTREAL &
    		    $86.7 {\rpm} 0.1$ &
    		    \underline{$85.7 {\rpm} 0.3$} &
    		    \underline{$65.7 {\rpm} 0.5$} \\
    		Diao 2021 &
    		    \underline{$87.1 {\rpm} 0.4$} &
    		    $81.2 {\rpm} 0.5$ &
    		    $52.2 {\rpm} 0.6$ \\
    		Caldas 2018 &
    		    $86.9 {\rpm} 0.3$ &
    		    $84.2 {\rpm} 0.3$ &
    		    $65.3 {\rpm} 0.5$ \\
    		\bottomrule
    	\end{tabular}%
        \caption{Final accuracy (after 7,500 rounds)}
        \label{fig:results_fl:final}
    \end{subfigure}
    \caption{
        Convergence during \ac{FL} on heterogeneous devices. \ourtech improves the convergence speed, while still reaching the same or a higher final accuracy than others.
	}
    \label{fig:results_fl}
\end{figure}

\paragraph{Heterogeneity Across Devices}
\label{sec:experiments:devices}

We first study heterogeneity across devices, i.e., devices have different resource availability but for now, there are no changes over time.
We select the resource availability at each device randomly and uniformly from a range with the upper bound being selected such that training the full \ac{NN} without dropout is possible.
The variability in the resource availability, i.e., ratio of
upper to lower bound
in the range, is reported in \cref{tab:fl_setting}.
We repeat every experiment three times and report the average and standard deviations of the classification accuracy.
\begin{figure*}[t]
    \centering
    \begin{subfigure}{\linewidth}
        \centering
	    \includegraphics{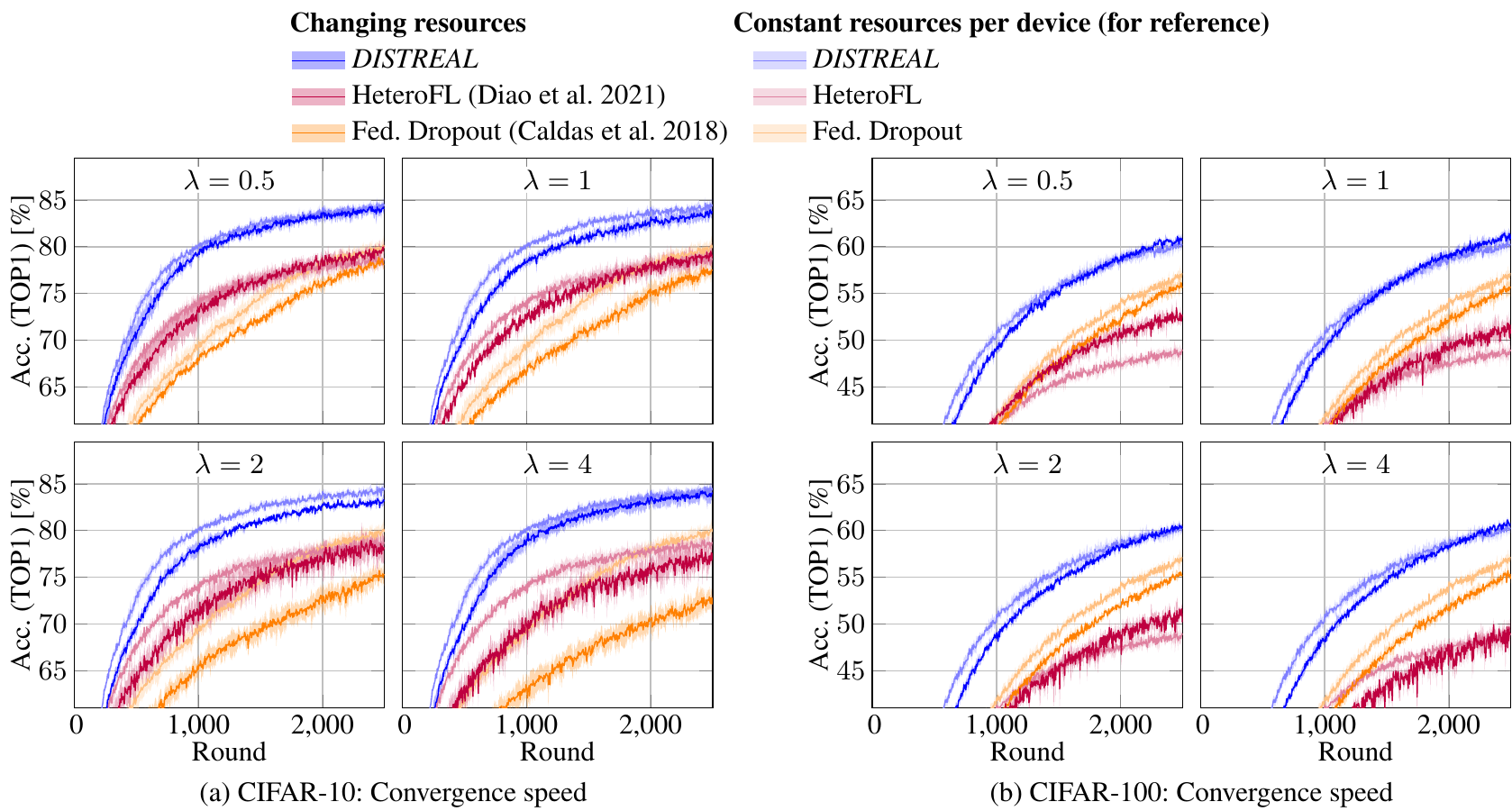}
    \end{subfigure}
    \begin{subfigure}{0.49\linewidth}
    	\centering
        \small
    	\vspace{2mm}
    	\begin{tabular}{ccccc}
    		\toprule
    		& $\lambda=0.5$ & $\lambda=1$ & $\lambda=2$ & $\lambda=4$\\
            \midrule
    		DISTREAL & 
    		    \underline{$86.3 {\rpm} 0.3$} &
    		    \underline{$86.0 {\rpm} 0.2$} & 
    		    \underline{$85.1 {\rpm} 0.3$} &
    		    \underline{$85.9 {\rpm} 0.4$} \\
    		Diao\,2021 & 
    		    $82.6 {\rpm} 0.2$ &
    		    $81.9 {\rpm} 1.3$ &
    		    $81.7 {\rpm} 0.5$ &
    		    $81.3 {\rpm} 0.3$ \\
    		Caldas\,2018 & 
    		    $83.4 {\rpm} 0.6$ &
    		    $83.2 {\rpm} 0.4$ &
    		    $82.4 {\rpm} 0.4$ &
    		    $82.7 {\rpm} 0.6$ \\
    		\bottomrule
    	\end{tabular}

    	\vspace{2mm}
        (c) CIFAR-10: Final accuracy (after 7,500 rounds)
    \end{subfigure}
    \hfill
    \begin{subfigure}{0.49\linewidth}
    	\centering
        \small
    	\vspace{2mm}
    	\begin{tabular}{ccccc}
    		\toprule
    		& $\lambda=0.5$ & $\lambda=1$ & $\lambda=2$ & $\lambda=4$\\
            \midrule
    		DISTREAL & 
                \underline{$65.4 \rpm 0.4$} &
                \underline{$65.5 \rpm 0.9$} &
                \underline{$64.7 \rpm 0.3$} &
                \underline{$64.7 \rpm 0.3$} \\
    		Diao\,2021 & 
                $57.2 \rpm 0.4$ &
                $56.8 \rpm 1.0$ &
                $56.5 \rpm 1.0$ &
                $56.6 \rpm 1.0$ \\
    		Caldas\,2018 & 
                $65.0 \rpm 0.3$ &
                $64.5 \rpm 0.2$ &
                $64.5 \rpm 0.3$ &
                $64.5 \rpm 0.3$ \\
    		\bottomrule
    	\end{tabular}

    	\vspace{2mm}
        (d) CIFAR-100: Final accuracy (after 7,500 rounds)
    \end{subfigure}
    \caption{Convergence with CIFAR-10 and CIFAR-100 on heterogeneous devices where resources availability changes randomly over the time with varying rate parameter~$\lambda$.
    \ourtech achieves a higher convergence speed than the state of the art, and the highest final accuracy, independently of $\lambda$.
    Experiments with full resource availability or with a small network are not repeated as they perform the same as in \cref{fig:results_fl}.
    }
    \label{fig:results_varying_fl}
\end{figure*}%

\cref{fig:results_fl} shows the accuracy results for the three datasets.
FEMNIST uses a simple network.
We observe that the small network baseline has the lowest convergence speed, with all other techniques showing similar performance.
The varying quantity (see \cref{tab:fl_setting}) and distribution of local data on the devices make the training noisy.
Nevertheless, \ourtech is not more sensitive to non-iid data than other solutions and reaches the same convergence speed and final accuracy.
With CIFAR-10, \ourtech achieves a significantly higher convergence speed than Federated Dropout or HeteroFL and reaches the accuracy of the theoretical baseline after 2,000 rounds.
The simple baseline that uses a smaller network on all devices initially converges faster but saturates early.
We also evaluate the final accuracy, where we train with each technique for 7,500 rounds. This ensures that all techniques have fully converged.
\ourtech and Federated Dropout reach the highest final accuracy, similar to the upper bound of full resource availability.
Similar observations can be made with CIFAR-100.

As the resources are not changing over time, the main contributions
of \ourtech
in this scenario are the application of the \ac{DSE}, which enable devices to efficiently utilize the available resources and the fact that \ourtech applies a probabilistic approach and drops different filters in different mini-batches, allowing to support a large number of resource levels.
It, therefore, outperforms Federated Dropout in terms of convergence speed, which uses the same dropout rates over all the layers, and HeteroFL, which supports only a few resource levels and removes the filters in always the same order, as \ourtech fully
utilizes the available resources, and uses these resources more efficiently w.r.t.\ convergence.
Besides, deeper is the trained model, higher is the relative gain.

\paragraph{Heterogenity Across Devices and Over Time}
\label{sec:experiments:both}

This section studies a fully heterogeneous \ac{FL} system, i.e., resource availability varies between devices and for each device over time.
As discussed in the introduction, this for instance due to shared resource contention between the learning task and other tasks.
The available resources for learning may change at any time (workload changes happen in the order of seconds \cite{tu2014performance}), i.e., also in the middle of a round.
As these changes stem from changes in the environment, they may be unpredictable to the device~\cite{duc2019machine}.
We model them as random, %
with the time between changes following an exponential distribution with rate parameter~$\lambda$.
The absolute resource availability levels are sampled from the same range as in the previous section, i.e., also according to \cref{tab:fl_setting}.
Thereby, the average resource availability across all devices and over time is the same as in the previous section.
We study four different values of  $\lambda\in\{0.5,1,2,4\}$, to simulate a range of slowly to rapidly changing scenarios. 

Figs.\ \ref{fig:results_varying_fl}a and \ref{fig:results_varying_fl}b show the convergence speed for CIFAR-10 and CIFAR-100, respectively.
We also plot the convergence speed with constant resources (previous results from \cref{fig:results_fl:convergence}) for reference.
We observe that for both datasets, the convergence speed with \ourtech is not dependent on the rate of resource changes and almost matches the results of the previous section.
In contrast, the convergence speeds of HeteroFL and Federated Dropout significantly degrade with higher~$\lambda$.
In addition, \ourtech reaches the highest final accuracy (Figs.\ \ref{fig:results_varying_fl}c and \ref{fig:results_varying_fl}d), independently of~$\lambda$.
The baselines with full resource availability and 
small model perform the same as in~\cref{fig:results_fl}, therefore are not shown again.

HeteroFL and Federated Dropout both select the trained model subsets on devices at the server at the start of a round.
The devices train on the assigned subset for the whole round and hence cannot react to potential unpredictable changes in the resource availability during a round.
An increase in resource availability results in underutilization of available resources, as training finishes early and the device is idle until the end of the round.
A decrease in resource availability results in the training not finishing in time (i.e., the device becomes a straggler), leading to the device being dropped from the round.
In contrast, \ourtech adjusts the resource requirements of training at run time by selecting a different dropout vector when a change occurs, finishing the training in time and fully utilizing the available resources.

These results show the importance of tackling the challenges discussed in the introduction, to \emph{fully} and \emph{efficiently} utilize available resources on each device.
Our technique achieves this by performing dropout at the devices, enabling them to react fast to the changes in a fine-grained manner.
This enables to fully utilize all available resources, making convergence robust to changes in the resource availability.
Furthermore, the \ac{DSE} enables us to efficiently utilize the available resources by finding Pareto-optimal dropout vectors w.r.t.\ resource requirements and achieved convergence speed.
These gains in convergence speed do not come at the cost of a lower final accuracy.

\section{Conclusion}
\label{sec:conclusion}

We addressed the problem of distributed training of \acp{NN} under heterogeneous resource availability, proposing \ourtech.
Our results show that \ourtech significantly improves the convergence speed in a heterogeneous \ac{FL} system, where resources vary both between devices \emph{and over time} without compromising on the final accuracy.
This is as our solution provides each device the capability to adjust the training in a fine-grained manner, enabling it to fully and efficiently utilize its available, but limited, resources.

\section*{Acknowledgments}
This work is in parts funded by the Deutsches Bundesministerium für Bildung und Forschung (BMBF, Federal Ministry of Education and Research in Germany).

\bibliography{bibliography}

\clearpage
\appendix

\section{Structured Dropout}

This section provides more details of filter-based structured dropout that is used to reduce the required resources (amount of computation) for training.

\subsection{Number of MACs}
\label{sec:macs}

Structured dropout is a probabilistic process, i.e., a different number of filters may be dropped in every mini-batch.
\cref{tab:macs} shows for the most important layers in \iac{CNN} how the \emph{expected} number of \acp{MAC} in the forward pass depends on the dropout rate.
For a convolutional layer, it depends not only on the dropout rate of this layer but also on the dropout rate of the preceding convolutional layer because a reduced number of filters (dropout of this layer) is applied to a reduced number of input feature maps (previous layer).
In the specific case that all per-layer dropout rates are the same, the expected number of \acp{MAC} reduces quadratically with the dropout rate.
Most other layers (batch normalization, activation, pooling) operate on each feature map independently.
Their expected number of \acp{MAC} scales linearly with the dropout rate.

\begin{table}[b!]
    \centering
    \begin{tabular}{crl}
        \toprule
        Layer Type & \multicolumn{2}{c}{MACs} \\
        \midrule
        Convolution & \multicolumn{2}{l}{$(1{-}d) {\cdot} |Y| {\cdot} \left((1{-}d_p) {\cdot} c_i {\cdot} k_w {\cdot} k_h + b\right)$} \\
        & \multicolumn{2}{l}{where} \\
        & $d$ &  dropout rate in this layer \\
        & $|Y|$ & nb.\ of output pixels \\
        & $d_p$ & dropout rate in the prev.\ layer \\
        & $c_i$ & input channels \\
        & $k_w$,$k_h$ & kernel width and height \\
        & $b$ & $1$ if bias is used, $0$ otherwise \\[2ex]
        Batch Norm., & \multicolumn{2}{l}{$(1-d_p)\cdot x$} \\
        Activation, & \multicolumn{2}{l}{where} \\
        Pooling & $d_p$ &  dropout rate in the prev.\ layer \\
        & $x$ & nb.\ of MACs w/o dropout \\
        \bottomrule
    \end{tabular}
    \caption{Expected number of MACs of the forward pass of individual layers with structured dropout.}
    \label{tab:macs}
\end{table}

\subsection{Resource Requirements and Availability}

Structured dropout reduces the theoretical amount of computation, i.e., the number of \acp{MAC}.
This also affects the resource requirements of training, which can be measured in many ways: execution time, energy, etc.
We study in this section the relationship between the theoretical amount of computation and the resource requirements and show that
\begin{enumerate}
    \item \acp{MAC} are a correct abstraction, i.e., they strongly correlate with the underlying resource.
    \item \acp{MAC} are a practical abstraction, i.e., it is possible to determine the available \acp{MAC}/s at run time.
\end{enumerate}
We pick the execution time of training on the CPU of a Raspberry Pi 4 with 2\,GB RAM as an example.
The implementation is done in 32-bit PyTorch~1.7.1.

\begin{figure}
    \centering
	\includegraphics{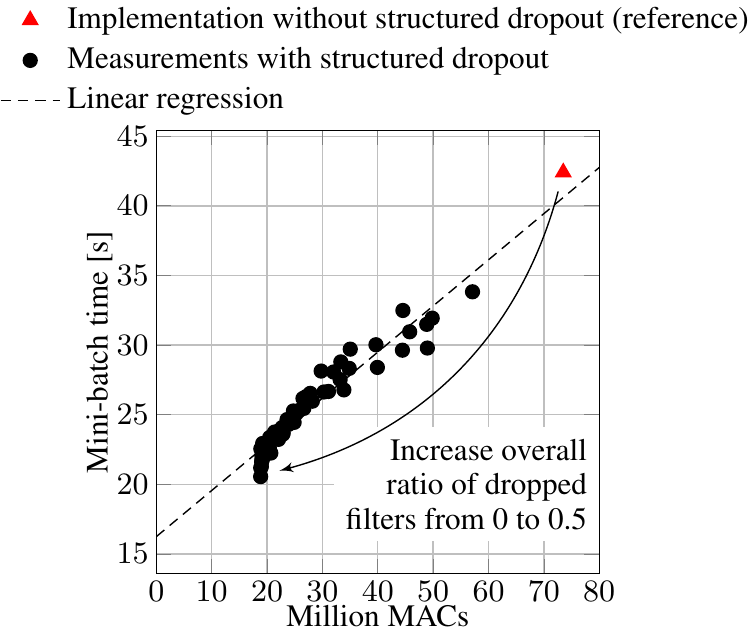}
    \caption{Mini-batch training time of DenseNet-40 for different dropout vectors on a Raspberry Pi 4 shows a linear relationship with the number of MACs in the forward pass.}
    \label{fig:macs_vs_time}
\end{figure}

For the first point, we measure the execution time of one mini-batch (size 128) of training DenseNet-40 with each dropout vectors found by the \ac{DSE}.
\cref{fig:macs_vs_time} plots the execution time (forward and backward pass) over the number of \acp{MAC} in the forward pass.
We make two main observations.
Firstly, the execution time is significantly reduced compared to an implementation of the same \ac{NN} without any dropout, i.e., with vanilla PyTorch layers.
This demonstrates the suitability to employ structured dropout to reduce the resource requirements (CPU time).
We observe that the copy overheads depend on the underlying architecture.
On platforms with separate VRAM for the GPU, copy overheads are higher.
An implementation revision is required to obtain similar benefits as on platforms with shared memory between CPU and GPU.
Because we target distributed devices, such as IoT devices that have a shared memory for CPU and GPU or even perform training on the CPU, this is a topic of future work.
Secondly, the mini-batch training time shows a clear linear relationship with the number of \acp{MAC}.
Consequently, the number of \acp{MAC} in the forward pass is a good approximation of the actual resource requirements during training, and can be used in the \ac{DSE} to determine the fitness of a dropout vector.
A constant offset has no impact, as the fitness is normalized to~$[0,1]$.

Note that in a general case, estimating the resource requirements from the total number of \ac{MAC} may be inaccurate~\citep{resources_are_complicated}.
This comes from the fact that different layer types may result in a different relationship between resource requirements and the theoretical number of \acp{MAC}.
This is for example the case if different layers cause different data movement between the main memory and the CPU, or if different layers can be accelerated differently by vectorization.
However, \emph{we use the number of \acp{MAC} only to compare different configurations (dropout vectors) of the same topology, i.e., we do not compare different topologies.}
Therefore, the number of \acp{MAC} is closely related to the resource requirements.

Due to this close correlation with the training time, the number of available \acp{MAC} for training can also be derived at run time.
As an example, we consider the CPU time as the limited resource due to resource contention.
In this example, the FL training is one of several tasks that are executed on the system, and that compete for CPU time.
The operating system scheduler decides the time during which each application is executed on the system based on many factors, such as fairness or priorities.
The amount of CPU time available for training is, therefore, known at the OS level and is made available to the application.
From the CPU time and desired throughput, one can directly obtain the time per mini-batch.
Finally, the number of \acp{MAC} can be derived from the mini-batch training time as in \cref{fig:macs_vs_time}.
This shows the practicality of such abstraction, proving the second point.

\subsection{Implementation in PyTorch}
\label{sec:pytorch}

This section summarizes our implementation of dropout in PyTorch, which is publicly available\footnote{\url{https://git.scc.kit.edu/CES/DISTREAL}}. %

\textbf{Convolutional layers~~}
A convolutional layer performs the following computation:
\begin{equation}
    O = I * W+B,
\end{equation}
where $O \in {\rm I\!R}^{b \times c_o \times w_o \times h_o}$ is the output activation ($b$ samples per mini-batch, $c_o$ output feature maps, width $w_o$ and height $h_o$ of the output feature maps), $I \in {\rm I\!R}^{b \times c_i \times w_i \times h_i}$ is the input activation ($c_i$ input feature maps, width $w_i$ and height $h_i$ of the input feature maps), $W \in {\rm I\!R}^{c_o \times c_i \times k_w \times k_h}$ (width~$k_w$ and height~$k_h$ of the convolutional kernel) and $B \in {\rm I\!R}^{c_o}$ are the weights and bias.

Structured dropout drops some of the filters from the computation, i.e., some of the feature maps in~$O$ are replaced by~$0$.
Dropout in a preceding layer results in some of the input feature maps to contain only~$0$.
To benefit from dropout in terms of computational requirements, unnecessary computations need to be avoided.
This is achieved by excluding dropped parts from the computations by performing sparse convolution.

\begin{figure}
    \centering
    \begin{subfigure}[b]{0.49\linewidth}
        \centering
		\includegraphics{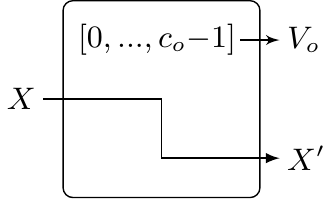}
    	\caption{Begin sparse part}
        \label{fig:implementation:begin}
    \end{subfigure}
    \begin{subfigure}[b]{0.49\linewidth}
        \centering
		\includegraphics{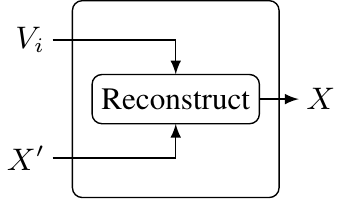}
    	\caption{End sparse part}
        \label{fig:implementation:end}
    \end{subfigure}
    \begin{subfigure}[b]{0.49\linewidth}
        \centering
		\includegraphics{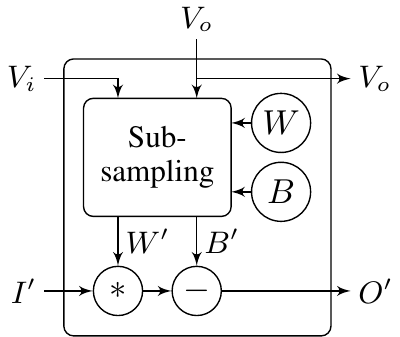}
    	\caption{Sparse convolution}
        \label{fig:implementation:conv}
    \end{subfigure}
    \begin{subfigure}[b]{0.49\linewidth}
        \centering
		\includegraphics{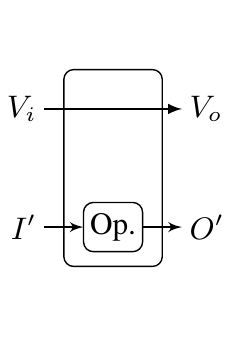}
    	\caption{ReLu, pooling, etc.}
        \label{fig:implementation:relu}
    \end{subfigure}
    \caption{Building blocks for sparse \ac{NN} that results from filter-based structured dropout.}
    \label{fig:implementation}
\end{figure}

Let~$V_i$ denote the set of the indices of valid (not dropped) feature maps in the input~$I$ of the convolutional layer.
$V_o$ is the set of the indices of valid feature maps in the output~$O$ and is obtained from a Bernoulli process with the dropout rate~$d$.
$O' \in {\rm I\!R}^{b \times |V_o| \times w_o \times h_o}$ represents the valid output feature maps in $O$, $I' \in {\rm I\!R}^{b \times |V_i| \times w_i \times h_i}$ represents the valid output feature maps in $I$.
It follows that
\begin{equation}
    O' = I' * W[V_o,V_i,:,:] + B[V_o],
\end{equation}
where $W[V_o,V_i,:,:]$ represents the parts of $W$ that are obtained by keeping only valid columns/rows according to $V_o$ and $V_i$ (same for $B[V_o]$).
The sparse convolution can be replaced by a regular (dense) convolution on the subsampled inputs.
This subsampling of $W$ and $B$ needs to be done at run time in every mini-batch because $V_i$ and $V_o$ may change in every mini-batch.
However, if the following layer is aware of $V_o$, the output~$O$ does not need to be reconstructed from $O'$, and $O'$ can be forwarded to the next layer unchanged.
Our implementation avoids copying input and output activations.
This is achieved by operating on sparse representations of $O$ and $I$ instead.
These are tuples $(O'$, $V_o)$ and $(I'$, $V_i)$.
However, this requires implementing a custom convolutional layer to manage the sparse computations.
\cref{fig:implementation:conv} depicts a sparse convolutional layer for structured dropout with rate~$d$.

PyTorch stores weight and activations in tensors, i.e., multi-dimensional matrices.
Tensors in PyTorch are regular, i.e., they store data in memory with fixed strides.
Subsampling the weights $W$ and $B$ according to $V_i$ and $V_o$ would results in irregular stride pattern, and therefore requires copying the underlying data in memory.
This could be avoided by extending the backend to be aware of $V_i$ and $V_o$, and still providing the full~$W$ and $B$ to the backend.
However, PyTorch supports several backends, some of which (e.g., CUDA) comprise proprietary code, and cannot be modified easily.
It is also important to notice that the weight and bias tensors in convolutional layers are much smaller than the input and output activation, lowering the copy overheads.

\textbf{Dropout~~}
Traditional dropout performs two operations during training: it replaces dropped values by~$0$ and scales non-dropped values by $1/(1-d)$.
In our implementation of structured dropout, these two operations need to be separated.
The first operation is performed by passing the indices of feature maps to compute~$V_o$ to the preceding convolutional layer.
Scaling needs to be done at a later step, potentially with other layers (like batch normalization) in between.

\textbf{Other layers~~}
Many layers, such as activation, pooling, etc.\, operate on individual feature maps independently.
As a result, these operations can directly be applied to the valid feature maps~$I'$.
The indices of the valid feature maps of the output of this operation~$V_o$ are the same as of the input~$V_i$, as depicted in~\cref{fig:implementation:relu}.
We require two additional layers that begin and end the sparse part of the \ac{NN}, respectively.
The begin layer (\cref{fig:implementation:begin}) annotates the activation with the information that all feature maps are valid ($V_o=[1,\ldots,c_o-1]$).
This does not require any copy.
The end layer (\cref{fig:implementation:end}) fills in zeros at the positions of invalid feature maps.
This requires copying tensors.
However, this layer is only required once.
We did not implement sparse fully-connected layers but similar concepts as for the convolutional layers can be applied.
\begin{figure}[t!]
    \centering
	\includegraphics{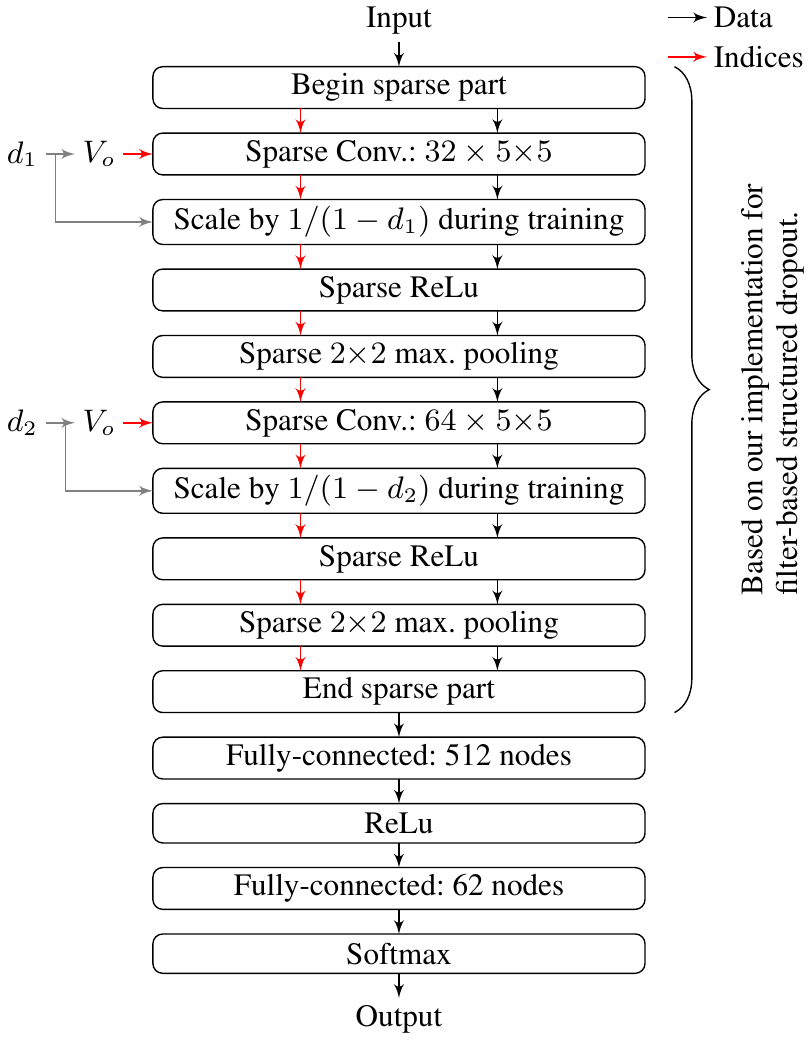}
    \caption{\ac{NN} for FEMNIST with sparse computations resulting from filter-based structured dropout.}
    \label{fig:feminst_network}
\end{figure}

\cref{fig:feminst_network} illustrates how these sparse layers can be used in a whole \ac{NN} that consists of a sparse convolutional part and a dense fully-connected part.

\section{Experimental Details}
\label{sec:setup_details}

\subsection{DSE}
\label{sec:dse_details}

This section provides details on how our \ac{DSE} employs the NSGA-II evolutionary algorithm from pygmo2~\citep{pagmo}.

\textbf{NSGA-II Parameters~~}
We use the default parameters provided in pygmo2, i.e., the crossover probability is set at 0.95, the distribution index for crossover is set at 10, the mutation probability is set at 0.01, and the distribution index for mutation is set at 50.
The population size is 64.
The size of individuals is 2, 39, and 99 continuous per-layer dropout rates for the networks used with FEMNIST, CIFAR-10 (DenseNet-40), and CIFAR-100 (DenseNet-100), respectively.
The minimum and maximum dropout rates are set at 0 and 0.5, respectively.

\textbf{Initial Population~~}
The initial population comprises 62 random individuals (dropout rates sampled randomly from $[0,0.5]$).
We add two individuals with dropout rates set at all 0 or all 0.5 to accelerate the exploration of the Pareto-front.

\textbf{Fitness Evaluation~~}
To evolve the population by one generation, first the fitness of each individual is assessed.
The resource requirements are the expected number of \acp{MAC} according to the equations in \cref{tab:macs}.
As explained in the section on resource-aware training, the convergence speed is measured by observing the accuracy change after short training.
This short training comprises 64 mini-batches of size 64.
The optimizer is \ac{SGD} with momentum set at 0.9 and weight decay of $10^{-4}$.
The learning rates are set at 0.005 for the network with FEMNIST, and 0.01 for DenseNet-40/100.
The training starts from a snapshot that has been pre-trained on a distorted dataset.
For CIFAR-10/100, this distortion changes the brightness, contrast, saturation and hue by 0.5 using PyTorch \texttt{ColorJitter} class.
For the grayscale FEMNIST, we rotate images by 90 degrees.
Training is repeated with three random seeds (used both for pre-training and short training) to obtain an average accuracy improvement.
Fitness values are normalized according to \cref{eq:fitness}.

\subsection{Miscellaneous Training Parameters}
\label{sec:fl_details}

This section lists parameters used in the \ac{FL} experiments that have not been included in the description of the experimental setup.
The local training on each device uses mini-batches of size 64.
The optimizer is the same as for the fitness evaluation (\ac{SGD} with momentum set at 0.9 and weight decay of $10^{-4}$).
The learning rates are set at 0.035 for the network with FEMNIST (as in \citep{caldas2018expanding}), and 0.05 for DenseNet-40/100.
The \ac{NN} for FEMNIST is also similar to the one used in \citep{caldas2018expanding}, i.e., \iac{CNN} with two $5{\times}5$ convolutional layers with 32 and 64 filters, respectively, each with ReLU activation and each followed by a $2{\times}2$ max pooling.
The convolutional part is followed by two fully connected layers with 512 and 62 neurons (number of classes), respectively.
The implementation of this \ac{NN} with structured dropout is depicted in \cref{fig:feminst_network}.
(The \acp{NN} for CIFAR-10/100 are explained in the description of the experimental setup.

\section{Compute Time for Experimental Results}
\label{sec:compute}

\begin{table*}
    \centering
    \begin{tabular}{cccccc}
        \toprule
         & & & \multicolumn{3}{c}{Total compute time} \\
        Dataset   & NN           & Generations & System 1 & or & System 2 \\
        \midrule
        FEMNIST   & Conv. NN     & 20 &  15\,h && 5.5\,h \\
        CIFAR-10  & Densenet-40  & 50 & 270\,h && 170\,h \\
        CIFAR-100 & Densenet-100 & 50 & 330\,h && 480\,h \\
        \bottomrule
    \end{tabular}
    \caption{Compute time to run the DSE (only once at design time).}
    \label{tab:compute_dse}
\end{table*}

\begin{table*}
    \centering
    \begin{tabular}{ccccccc}
        \toprule
        & & & & \multicolumn{3}{c}{Total compute time} \\
        Dataset & Resource heterogeneity & Rounds & Experiments  & System 1 & or & System 2 \\
        \midrule
        FEMNIST   & across devices          & 7,500  & 5${\times}$3 &   800\,h &&   500\,h \\
        CIFAR-10  & across devices          & 7,500  & 5${\times}$3 &   570\,h && 1,260\,h \\
        CIFAR-100 & across devices          & 7,500  & 5${\times}$3 & 1,170\,h && 4,290\,h \\
        CIFAR-10  & across devices and time & 7,500  & 3${\times}$4${\times}$3 & 1,470\,h && 3,150\,h \\
        \bottomrule
    \end{tabular}
    \caption{Compute time to reproduce the FL results.}
    \label{tab:compute_fl}
\end{table*}

We report in this section the compute time required to reproduce our experiments.
\emph{These are not the resources that we model on the devices}.
We run our experiments on two types of systems.
The first type of systems (System 1) runs Ubuntu~18.04.6~LTS and contains an Intel Core i5-4570 with 32\,GB RAM and a NVIDIA GeForce GTX~980 with 4\,GB VRAM.
The second type of systems (System 2) runs \mbox{CentOS}~7.9.2009 and contains an AMD Ryzen~7~2700X with 64\,GB RAM and no GPU.
We report the total compute time without considering parallelization to several machines.

\cref{tab:compute_dse} shows the compute time for our \ac{DSE}, which runs only once at design time.
Compute times report the total compute if run on \emph{either} of the two systems.
Different networks require different numbers of generations to converge.
We use three random seeds to measure the accuracy change after short training for a dropout vector.

\cref{tab:compute_fl} shows the compute times required to reproduce our experimental results with \ac{FL}.
The number of rounds with the three datasets is reported in \cref{tab:compute_fl}.
The experiments with heterogeneity across devices but not over time, each comparing five techniques, are repeated with three random seeds.
The experiments with heterogeneity across devices and over time, comparing three techniques in four different rates of change in the resource availability $\lambda$, are repeated with three random seeds.

\section{Computations in CNNs}

\begin{figure}
	\centering
	\includegraphics{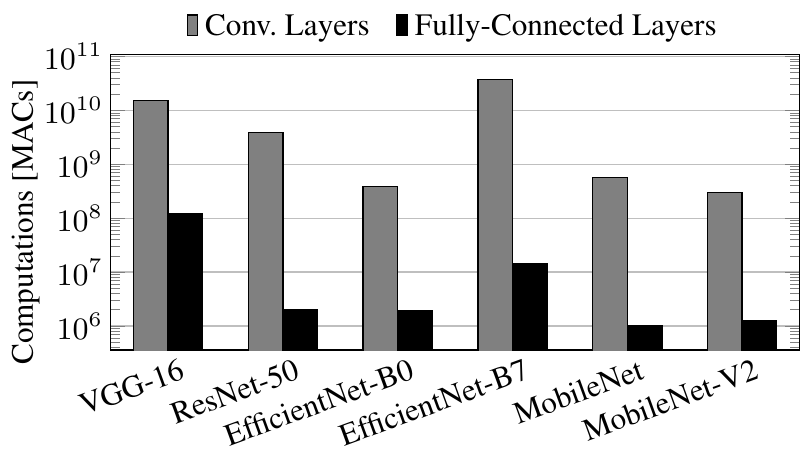}
	\caption{
		Convolutional layers account for the majority of MAC operations in \acp{CNN}.
	}
	\label{fig:conv_ops}
\end{figure}

\cref{fig:conv_ops} shows how the number of \acp{MAC} for the forward pass is distributed between fully-connected and convolutional layer types for different state-of-the-art \acp{NN}.
We observe that for all architectures, the convolutional part requires orders of magnitude more \acp{MAC} in total than the fully-connected part.

\end{document}